\documentclass[runningheads]{llncs}
\usepackage{graphicx}
\usepackage{amsmath} 
\usepackage{mathtools}
\usepackage{algorithm}
\usepackage[noend]{algpseudocode}
\usepackage{multirow}
\usepackage{tikz}
\usetikzlibrary{shapes.geometric, arrows}
\usetikzlibrary{fit,positioning,arrows}

\begin{document}
\title{Tracking in Urban Traffic Scenes from Background Subtraction and Object Detection}
\titlerunning{Tracking from Background Subtraction and Object Detection}
%
 \author{Hui-Lee Ooi \and
 Guillaume-Alexandre Bilodeau \and
 Nicolas Saunier}
\authorrunning{Ooi et al.}
%
\institute{Polytechnique Montr\'eal, Montr\'eal, Canada\\
\email{hui-lee.ooi@polymtl.ca, gabilodeau@polymtl.ca, nicolas.saunier@polymtl.ca}}
\maketitle              
\begin{abstract}
In this paper, we propose to combine detections from background subtraction and from a multiclass object detector for multiple object tracking (MOT) in urban traffic scenes. These objects are associated across frames using spatial, colour and class label information, and trajectory prediction is evaluated to yield the final MOT outputs. The proposed method was tested on the Urban tracker dataset and shows competitive performances compared to state-of-the-art approaches. Results show that the integration of different detection inputs remains a challenging task that greatly affects the MOT performance.



\keywords{Multiple object tracking  \and Urban traffic scene \and Road user detection.}
\end{abstract}

\section{Introduction}
The task of multiple object tracking (MOT) is to produce a set of trajectories that represent the actual real-life movements of the objects of interest across frames. In the context of urban scenes such as traffic intersection, MOT is performed for the road users (vehicles, pedestrians, cyclists, motorcyclists, etc.) as objects of interest for the purpose of traffic control and management to improve traffic while mitigating the adverse impacts. Due to the nature of such settings, interactions among the objects are expected and frequent, thus leading to object occlusions. Compared to conventional traffic scenes where the speeds of the road users are usually more consistent and directions homogeneous, MOT in urban traffic scenes remains a difficult and challenging task as it deals with objects interacting in different directions and speeds. Furthermore, because of the typical camera setups used, object scales varies significantly, which can make them difficult to detect. 

The advances and reported good results in recent years of multiclass object detection algorithms with deep learning \cite{girshick2016region} have prompted us to integrate them into the tracking process. In addition, the class label information can provide a useful description of objects to help with their association across frames in the tracking steps. However, the recent work of Ooi et al. \cite{ooi2018multiple} has shown that tracking with a multiclass object detector (MOD) is very challenging since detections are often incorrect or missing.  Since the incorrect or missing detection of an object at that stage can propagate and leave a huge impact on the final tracking results, we seek to improve the detection inputs in order to achieve better MOT results. Therefore, we extend the work of Ooi et al. \cite{ooi2018multiple} by using as inputs, detections from both a MOD and a background subtraction algorithm. To handle the problem of occlusion, a Kalman filter is used for prediction when an object of interest is not seen at the detection stage. This helps in keeping track of object of interest that might have been hidden by other objects at certain time steps during the lifespan of the trajectory. 


In this paper, we introduce a MOT solution for urban traffic scenes with fused inputs from the integration of background subtraction inputs \cite{beaupre2018improving} with detections from a pre-trained MOD. Our two main contributions are: 1) a novel method to fuse detections from two sources that may contradict each other and 2) an object descriptor based on object class labels and their learned detection confidence.

\section{Related Works}
MOT usually comprises several steps: 1) object detection, 2) appearance modeling, and 3) data association. A large part of the past literature on MOT emphasized the challenge of data association \cite{bewley2016simple} and its effect on MOT performance. Researchers proposed sophisticated data association strategies that often extend the Hungarian algorithm. For example, the Joint probabilistic data association filter (JPDAF) tracks objects based on the most likely outcome for each trajectory by considering every detection available, as well as missing or spurious detections \cite{Rezatofighi2015,CSM2009Shalom}. Another example is the minimum-cost flow algorithm that formulates the data association problem as finding the shortest path from the apparition of the object to its last appearance in the scene \cite{CVPR2011Pirsiavash}.

On the other hand, object detection is necessary before data association, as poor detections will severely deteriorate the tracking performance. Hence, some previous MOT solutions have proposed combining detection methods to allow better object inputs for improved tracking in the end. The main drawback of using inputs from background subtraction is the difficulty of distinguishing the merging, fragmentation and splitting of objects. In cases where multiple road users are in close proximity, partial occlusion will cause the incorrect merging of these road users. IMOT (Improved Multiple Object Tracking) was introduced by Beaupr\'e et~al. \cite{beaupre2018improving} as an improved version of background subtraction using edge processing and optical flow, converting blobs of objects into compact bounding boxes that outline individual objects if there is evidence based on motion that two or more objects were grouped together. 

More generally, the MOT problem in urban scenes was tackled several times in the past. A combination of background subtraction and feature points were proposed in Urban tracker \cite{jodoin2016tracking}. Based on detections from background subtraction, objects are described by several keypoints which provide robustness to partial occlusion as a subset of keypoints can be matched if they are not all hidden. MKCF \cite{yang2017multiple} was proposed as a solution for MOT, combining the background subtraction with multiple individual KCF (Kernelized Correlation Filters) single object trackers \cite{henriques2015high}. This method capitalizes on the robustness of newer visual object tracker. It shows good performances even if it uses rudimentary data association. Saunier et~al. \cite{saunier2006feature} used optical flow to detect the motion of objects of interest and the classic  Kanade-Lucas-Tomasi (KLT) framework \cite{Shi1994} to match road users from frame to frame. Recently, Ooi et~al. \cite{ooi2018multiple} used a MOD for road user tracking. However, the tracking performance was severely impacted by the inadequate and inconsistent detections across frames. 


\section{Methods}
\begin{figure}
\centering
\tikzstyle{inputbox} = [rectangle, rounded corners, minimum width=0.5cm, minimum height=1cm,text centered, text width=1.8cm, draw=black, fill=none]
\tikzstyle{computebox} = [rectangle, minimum width=0.5cm, minimum height=1cm,text centered, text width=1.8cm, draw=black, fill=none]
\tikzstyle{arrow} = [thick,->,>=stealth]
\tikzstyle{bidir}= [thick,<->,>=stealth]

\begin{tikzpicture}[node distance=2cm]
\node (det) [inputbox] {MOD objects};
\node (imot) [inputbox, below of=det, yshift=-0.5cm] {IMOT objects};
\node (fusion) [computebox, right of=det, xshift=0.1cm, yshift=-1.2cm, minimum height = 1.5cm] {Object fusion};
\node (description) [computebox, right of=fusion, xshift=0.5cm, minimum height = 1.5cm] {Object description};
\node (association) [computebox, right of=description, xshift=0.5cm, minimum height = 1.5cm] {Data association};
\node (tracking) [inputbox, right of=association, xshift=0.5cm, minimum height = 1.0cm] {Tracks};
\draw [arrow] (det) |- (fusion);
\draw [arrow] (det) -| (description);
\draw [arrow] (imot) -| (fusion);
\draw [arrow] (fusion) -- (description);
\draw [arrow] (description) -- (association);
\draw [arrow] (association) -- (tracking);
\node (Unmatched_t) [inputbox, below of=association] {Unmatched tracks};
\node (Unmatched_d) [inputbox, right of=Unmatched_t, xshift=0.5cm] {Unmatched detections};
\node (Matched_d) [inputbox, left of=Unmatched_t, xshift=-0.5cm] {Matched detections};
\draw [bidir] (association) -- (Unmatched_t);
\draw [bidir] (association) -- (Unmatched_d);
\draw [bidir] (association) -- (Matched_d);


\node(fitbox)[draw,dashed,text badly centered,fit={(Matched_d) (Unmatched_t) (Unmatched_d)}] {};
\end{tikzpicture}
\caption{Overview of our tracking framework. Object detections from two methods are first fused. They are described and associated across frames using sets of matched and unmatched tracks and detections. Based on these, the final tracks are outputted. } 
\label{fig_flowchart}
\end{figure}
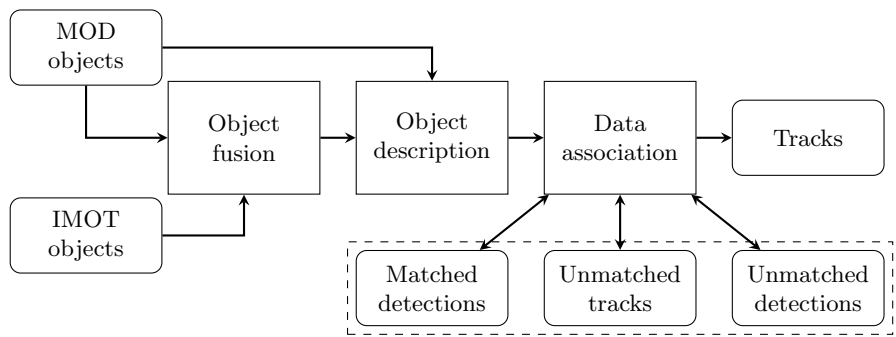

Three steps are involved in the proposed MOT strategy: (i)~Fusion of objects from detection methods, (ii)~Object description, and (iii)~The association of objects across frames. Our proposed method is illustrated in Figure~\ref{fig_flowchart}. It starts by fusing the input from a multiclass object detector (MOD) and from the improved background subtraction method IMOT. The resulting object detections are then tracked. Objects are described using colour, position and the class labels coming from the MOD. Then, data association is performed. 

\subsection{Object fusion}
In our proposed method, we integrate the bounding boxes from both IMOT and MOD into our tracking framework. The MOD objects are the result of the application of a pre-trained deep learning detection network, in our case RFCN~\cite{girshick2016region}, that was fine-tuned on the MIO-TCD dataset \cite{luo2018mio} containing varied road users such as cars, buses, bicycles, pedestrians, pickup trucks, etc. IMOT objects are the results of a post-processing over a background subtraction method in order to separate erroneously merged road users using edges and optical flow \cite{beaupre2018improving}. 

The objects from the two sources are matched and filtered before starting the tracking process. Due to the nature of IMOT objects, there could be some small bounding boxes that are not relevant as a result of shaking cameras and moving background elements. On the other hand, MOD objects will include long-term stationary road users that are beyond the scope of interest of our applications and there are occasions where objects of interest are missed out \cite{ooi2018multiple}. We hypothesize that the merged inputs can be fed into the tracker to give more satisfactory tracking results. However, results are often contradictory, sometimes IMOT gives better results while sometimes, it is MOD. One cannot simply merge the two sets of detections. 

The following fusion strategy is proposed. We assume that all IMOT objects are relevant to the tracking framework as stray small IMOT objects that are not representative of objects of interest are filtered out according to size prior to the matching. IMOT objects were shown to be more reliable. They also had better performance in detecting the objects in the scene. MOD objects are used to provide class labels and to merge fragmented IMOT objects. 

For matching the input objects of both sources, we compared their similarities in terms of bounding box (BB) overlaps and colour histogram. The BB overlap $S_o$ is given by 


\begin{equation}
S_{o} = \frac{A_i\cap B_j}{A_i\cup B_j},
\label{eqn:sim_overlap}
\end{equation}
where $A_i$ denotes the $i^{th}$  BB from IMOT output whereas $B_j$ denotes the $j^{th}$ BB from MOD output. We also calculate the colour similarity between IMOT objects and between IMOT and MOD objects. The colour similarity $S_c $ is given by

\begin{equation}
S_{c} =\sqrt[]{1-\frac{1}{\sqrt{\bar{G}\bar{H}N^2}}\sum_{i=1}^{N}{\sqrt{G_i H_i}}}, 
\label{eqn:sim_clr}
\end{equation}
where $G$ denotes the colour histogram of a first BB and $H$ denotes the colour histogram of a second BB. $N$ is the total number of histogram bins. $\bar{G}$ and $\bar{H}$ are the mean of the $N$ bins.

Pairings between IMOT objects and MOD objects are performed based on the overlap of the BBs with Equation \ref{eqn:sim_overlap} and a threshold $T_o$. IMOT objects that are matched with MOD objects will benefit from the class label information of MOD objects for data association in the tracking phase. On the contrary, IMOT objects that do not matched with MOD objects will be fed into the tracker with a dummy class label. There are cases where the matching is not one-to-one. For instance, several IMOT objects could be matched with the same MOD object and vice versa, though the latter is a rare occurrence since IMOT objects are usually smaller in size and more compact. MOD objects, on the other hand, are larger and often encompassed several objects at the same time. Hence, the merging of input objects is performed only on the IMOT objects and only if the colour of the objects to merge are similar enough based on Equation~\ref{eqn:sim_clr}. 

\begin{figure}[ht!]
\centering
\includegraphics[width=0.35\linewidth, trim={5cm 10.8cm 20cm 0},clip]{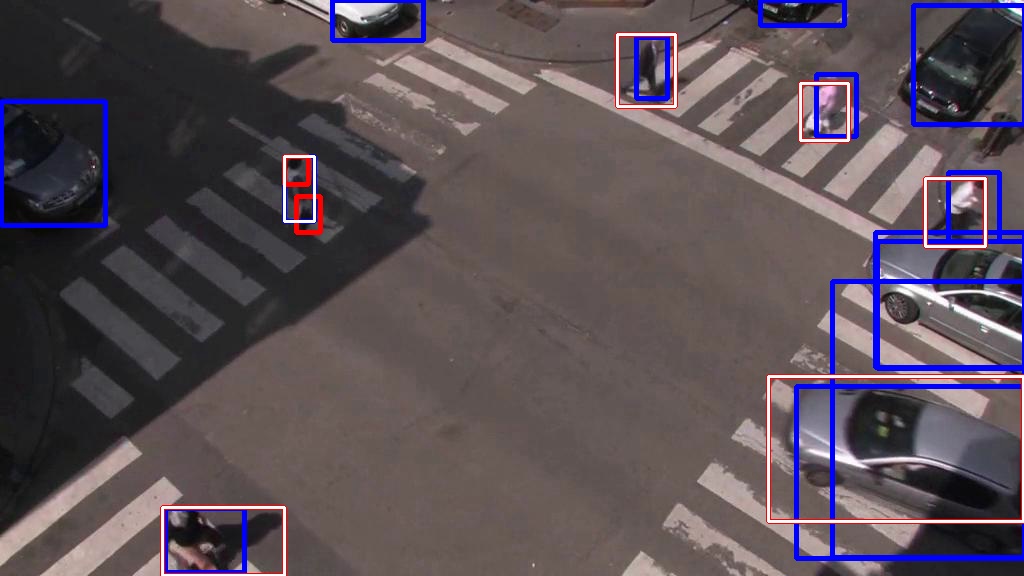}
\caption{Example of the merging of objects. Blue BBs: MOD objects, red BBs: IMOT objects, white BB: resulting fusion of the two inputs into the whole object (pedestrian).} 
\label{fig_fuse}
\end{figure}

Algorithm~\ref{code_fusion} describes the process for obtaining the final fused inputs for the MOT task. If multiple IMOT objects are matched to a particular MOD object, the colour histogram similarities of the multiple IMOT objects will be compared among themselves using the Bhattacharyya distance (Equation~\ref{eqn:sim_clr}) and a threshold $T_c$. If the similarity is high, then these objects are thought to be fragmented parts of an object and hence the BB from the MOD object would be taken as the input for the tracker. If the similarity is low, these objects will be considered individually in the subsequent steps. Figure~\ref{fig_fuse} illustrates how fragmented parts of an object are fused together to recover the whole object after taking into consideration the output of the MOD.

Colour similarity is used for merging IMOT because there are cases where the BB from the MOD object contains more than one actual object that should not be merged. Hence, care must taken to handle the different cases. To avoid excessive merging of IMOT objects that overlap with often large MOD objects, merging IMOT based on pairings between objects from the two approaches will only be evaluated if there is significant overlap (larger than $T_m)$. Finally, when a single IMOT object is matched with multiple MOD objects, the similarity in terms of BB overlap and colour histogram will be used to determine the final label from MOD that will be used with the IMOT object. 
\algnewcommand\algorithmicforeach{\textbf{for each}}
\algdef{S}[FOR]{ForEach}[1]{\algorithmicforeach\ #1\ \algorithmicdo}
\begin{algorithm}[h]
\caption{IMOT and MOD object fusion}
\label{code_fusion}
\begin{algorithmic}[1]
\Procedure{IMOT-MOD Pairing}{}
\ForEach{\texttt{IMOT objects}}
\ForEach{\texttt{MOD objects}}
\State 	Compute overlap of BBs with Eq. \ref{eqn:sim_overlap}  
\If {$S_{o} >= T_{o}$} 
\State Assign as pairs and update pairing matrix
\EndIf
\EndFor    
\EndFor
\EndProcedure

\Procedure{Merging multiple IMOT into single detection object}{}
\ForEach{\texttt{MOD objects}}
\If {\texttt{Pair with more than one IMOT}}
\ForEach{\texttt{IMOT objects that are paired}}
\State Compute colour similarity with Eq. \ref{eqn:sim_clr}
\If{$S_o >= T_{m}$ and  ${S_{c}}<= T_{c}$ }
\State Use MOD object as tracker input, discard the IMOT object
\Else  { Keep IMOT object}
\EndIf
\EndFor
\EndIf
\EndFor
\EndProcedure

\Procedure{Update IMOT object with label from paired MOD object}{}
\ForEach{\texttt{Remaining IMOT objects}}
\If{No pairing found}
\State Use the IMOT as tracker input with dummy label
\ElsIf {One-to-one pairing found}
\State Use the IMOT as tracker input with label from paired MOD
\Else
\ForEach{\texttt{Paired MOD objects}}
\State Compute $S_c$ and $S_o$ of IMOT with each MOD object
\State Use IMOT as input with label of MOD object with largest similarity
\EndFor
\EndIf
\EndFor
\EndProcedure
\end{algorithmic}
\end{algorithm}


\subsection{Data association costs}
 The cost of assigning pairings among the objects across frames is calculated by using the Hungarian algorithm \cite{kuhn1955hungarian}. The cost of matching a detected object and a tracked object is in the range of 0 and 1. The lower the cost, the more likely the two objects are referring to the same object.

For matching the objects across frames, the spatial cost $C_d$ is measured by the spatial distance between BBs of the compared objects using 
\begin{equation}
C_{d} = 1 - max (0, \frac{T_{d} - \bar{SD}}{T_{d}}) 
\label{eqn:cost_spatial}
\end{equation}

\begin{multline}
\bar{SD} =\frac{1}{4}(|x_{D,min} - x_{T,min}| + |y_{D,min} - y_{T,min}| + \\ |x_{D,max} - x_{T,max}| + |y_{D,max} - y_{T,max}|),
\label{eqn:distance_avg}
\end{multline}
where $x_{min}$ and $y_{min}$ denotes the minimum x and y coordinates, whereas $x_{max}$ and $y_{max}$ denotes the maximum x and y coordinate of an object. $T$ indicates an object that is currently tracked and $D$ indicates a detected object in a frame. $\bar{SD}$ is the mean spatial distance of the $x$ coordinates and $y$ coordinates of the four corners of the BBs of the compared objects whereas a fixed parameter $T_{d}$ is used to penalize objects that are too far and to normalize $C_{s}$.


For describing objects in terms of appearance, colour cost $C_c$ is computed based on the Bhattacharyya distance on colour histogram as in Equation \ref{eqn:sim_clr}, 
where $G$ denotes the colour histogram of a detection and $H$ denotes the colour histogram of a currently tracked object. $N$ is the total number of histogram bins (we used 256).


Finally, the class labels are also considered in the matching cost. Detection confidence is used in our formulation. $C_l$ is given by

\begin{equation}
  C_{l}=
  \begin{cases}
    1 - 0.5\times(W_{i} + W_{j}) & \text{if $L_{i}=L_{j}$ } \\
    1 & \text{if $L_{i} \neq L_{j}$}, 
  \end{cases}
\label{eqn:cost_label}
\end{equation}
where $L_{i}$ denotes the class label of object $i$ and $W_{i}$ its confidence value (between 0 and 1). As we will see in the results, using the confidence value from the MOD, and not just the class label for the cost is a beneficial strategy since confidence values tend to be similar in consecutive frames for a given object. 

The final association cost is a combination of $C_{d}, C_{c}, C_{l}$, and is given by

\begin{equation}
  C_{final}= \alpha C_{d} + \beta C_{c} + \gamma C_{l},
\label{eqn:cost_final}
\end{equation}
where $\alpha, \beta, \gamma$ denotes the weights for the corresponding cost.

\subsection{Overall Tracking Framework}
In the tracking phase, each input object that appeared at the start of the video will be included into a set that contains all the active objects, thereafter denoted as the tracked objects. New input objects in the subsequent frames are denoted as detected objects and are matched accordingly to the tracked objects. We enforce one-to-one matching using the Hungarian algorithm \cite{kuhn1955hungarian}, since it is expected that there exists only one true object at the next frame that corresponds to a currently tracked object. In addition, because of the non-ideal cases caused by occlusions or objects missing from the inputs that are common in urban scenes, some predictions are used to compensate the shortcomings of the inputs. 

Hence, for each processed frame, sets of matched detections, unmatched detections and unmatched tracks are obtained. Matched detections are essentially the successful pairings of detected objects and tracked objects. Unmatched detections refers to detected objects without pairing with the existing set of tracked objects. This can be due to the entrance of a new object into the scene or as a result of spurious detections from the inputs. Unmatched tracks are when there is no corresponding pairing found in the set of detected objects. This is usually due to occlusion or being missed by IMOT, but also by objects that have left the scene. 

For each active tracked objects, a Kalman filter is used to get a prediction of its expected location in the subsequent frame based on its history. If the tracked objects are matched with the detected objects, the prediction result will be discarded and the tracked object will be updated with information from the latest matched detected object. In the case where a tracked object is unable to find a matching counterpart in the set of detected objects, the prediction result may be used instead if it is deemed good. For each step of a track, the state or quality of the tracking is defined as "D" (Detection), "GP" (Good Prediction), "BP" (Bad Prediction) or "UP" (Uncertain Prediction). Overlap between the prediction result and the previous position in the trajectory (history) is used to evaluate the quality of a prediction. If the previous time stamp is marked as "D" (indicating it is from a matched pairing) or "GP" (indicating it as a reasonably good prediction), there is a good chance that the trajectory history is reliable. Hence, if the overlap is high (larger than $T_{p}$) between the prediction at the current step and the previous history step, the prediction result will be used and the state will be marked as "GP". If the overlap is not good, the state will be marked as "BP" and the prediction result will not be used. Instead the previous result in the tracking history will be used. For cases of unmatched tracked objects with a history that is not marked "D" or GP, the state will be marked at "UP" since there is no known reliable history that can be used to verify the current prediction. Algorithm~\ref{code_prediction} summarizes the inspection of the tracking prediction quality. 

\begin{algorithm}
\caption{Checking prediction quality for unmatched tracks}
\label{code_prediction}
\begin{algorithmic}[1]
\Procedure{Prediction Quality}{}
\ForEach{\texttt{Unmatched track}}
\If{ Previous time step is "D" or "GP"}
\If{Overlap of prediction with previous time step $< T_{p}$ }
\State Use BB output from previous time step and mark as "BP" 
\Else
\State Use prediction and mark as "GP"
\EndIf
\Else
\State Use prediction but mark as "UP"
\EndIf
\EndFor
\EndProcedure
\end{algorithmic}
\end{algorithm}

At the end of the tracking process, trajectories with significant amount of "BP" and "UP" will be removed eventually since these final trajectories are likely to contain incorrect prediction that does not reflect the actual movement of the objects of interest. 
 
For active track management, when a tracked object is unable to find a matching detection object for $T_{n}$ frames, the object is assumed to have left the scene. The track will therefore be terminated along with its last $T_n$ steps of tracking results removed since they are most likely not valid.

\section{Experiments}
The proposed method was tested on the Urban tracker dataset~\cite{jodoin2016tracking} and compared with several state-of-the-art methods. We also performed an ablation study on the data association cost components. The dataset includes four videos: Rouen, Sherbrooke, St-Marc and Rene-Levesque. We chose this dataset because it includes a large variety of object classes and background subtraction is applicable. 

The tracking performance is evaluated by using the CLEAR MOT metrics \cite{milan2016mot16} that are comprised of MOTA (Multiple Object Tracking Accuracy) and MOTP (Multiple Object Tracking Precision). MOTA evaluates the tracking performance by taking into consideration the number of objects that are mismatched, the false positives (FP) and false negatives (FN). MOTP evaluates the quality of the localization of the matches by checking the similarity of true positives (TP) with the corresponding targets in ground truths. 

We also report the following information. Ground truth (GT) is the number of actual object instants in the whole video. Misses are missing GT object instances in tracks. FP are spurious object detections that are not in the ground truth. Mismatches are the number of tracks that suffer from object identity switches. The identification of correct tracks, misses and FP are based on the overlap of bounding boxes from our tracking output with respect to the ones of the ground truth. We used a threshold of 0.3 for the overlap in tracking performance evaluation as proposed by Beaupr\'e et~al. \cite{beaupre2018improving}. In our experiments, $T_o = 0.05 $, $T_m = 0.5 $, $T_c = 0.5 $, $T_d = 0.5 $ and $T_p = 0.01 $.

\subsection{Ablation study}
We start our evaluation of our method with an ablation study on the data association cost. The individual effects of the cost components are compared in Table \ref{tab_rouen}. Generally it is observed that the spatial cost has the smallest number of mismatches and FP for all the evaluated videos. Since the spatial cost is based on the proximity of BBs, it is an essential component that describes the similarity of objects to determine across frames. In the results for St-Marc and Rene-Levesque, it has the highest number of correct tracks compared to the other association costs. 

Colour cost gives slightly inferior tracking performance, having more FP and mismatches with slightly fewer correct tracks compared to the spatial cost. This could be due to presence of multiple objects that share similar colour properties and the fact that proximity is ignored. In addition, since BBs contain a certain portion of background as well (depending how well the object is enclosed in the BB), this might not be the best cost component. However, it can disambiguate the association of nearby objects with different colours.  

Lastly, the class label cost gives the lowest performing tracking results due to reasons similar to the colour cost. There could be several objects that share the same class label in the same frame. With only the class label information it is often insufficient to do the right pairings. Also, some IMOT objects are fed into the tracker with dummy class labels since they are not paired with MOD objects in the object fusion stage. Nevertheless, the performance with this feature is better than expected thanks to the similarity of confidence values for the same object between frames. Since, the confidence value is used in Equation~\ref{eqn:cost_label}, objects are both discriminated by their class and the confidence value. 

\begin{table}[h]
\centering
\caption{Comparison of individual association cost components for the four videos of the Urban tracker dataset. \textbf{Bolface} indicates best result.}
\label{tab_rouen}
\begin{tabular}{ c | c | c | c c c c c c}
\hline 
 & Cost & GT                    & Correct Tracks & Misses & FP  & Mismatches & MOTP   & MOTA   \\
\hline
Rouen &Distance & \multirow{3}{*}{2627} & 2125           & 502    & \textbf{519} & \textbf{19}         & \textbf{0.604}  & \textbf{0.604} \\
& Colour    &                       & 2126           & 501    & 560 & 28         & 0.603 & 0.586 \\
& Label    &                       & \textbf{2128}           & \textbf{499}    & 804 & 143        & \textbf{0.604} & 0.450 \\
\hline 
Sherbrooke & Distance & \multirow{3}{*}{4429} & 3029           & 1400   & \textbf{400} & \textbf{1}          & 0.582 & \textbf{0.593} \\
& Colour    &                       & \textbf{3030}           & \textbf{1399}   & 401 & 6          & 0.582 & 0.592 \\
& Label    &                       & 3006           & 1423   & 503 & 45         & \textbf{0.584} & 0.555  \\
\hline 
St-Marc & Distance & \multirow{3}{*}{8375} & \textbf{6068}           & \textbf{2307}   & \textbf{515}  & \textbf{73}         & 0.696 & \textbf{0.654} \\
& Colour    &                       & 6041           & 2334   & 591  & 93         & 0.696 & 0.640 \\
& Label    &                       & 5820           & 2555   & 1161 & 293        & \textbf{0.700} & 0.521  \\
\hline 
Rene-Levesque & Distance & \multirow{3}{*}{9418} & \textbf{2701}           & \textbf{6717}   & \textbf{530} & \textbf{0}          & 0.740 & \textbf{0.231} \\
& Colour    &                       & 2694           & 6724   & 538 & 15         & 0.741 & 0.227 \\
& Label    &                       & 2596           & 6822   & 687 & 80         & \textbf{0.746} & 0.194  \\
\hline
\end{tabular}
\vspace*{-\baselineskip}
\end{table}

\subsection{Comparison with state-of-the-art methods}
The performance of the proposed method is compared with previous state-of-the-art work, IMOT \cite{beaupre2018improving}, Urban tracker \cite{jodoin2016tracking}, MKCF \cite{yang2017multiple} and Ooi et~al. \cite{ooi2018multiple} that were evaluated on the Urban tracker dataset. For the data association cost, the weights of spatial, colour and label costs are 0.6, 0.3 and 0.1 respectively for $\alpha, \beta $ and $\gamma$. As shown in Table~\ref{tab_comparison}, the proposed method yields better tracking performance than Urban Tracker, MKCF and Ooi et al. \cite{ooi2018multiple}. Overall, IMOT outperformed all evaluated methods, even though our proposed method performs the best in terms of MOTA for the video St-Marc and is second best in terms of MOTA on Sherbrooke. It is noted, however, that the proposed method gives a low MOTA for Rene-Levesque. Fusion of objects in the proposed method is not working well for this particular video as the objects in the scene are very small, and inevitably they get incorrectly paired with MOD bounding boxes that are usually large and imprecise for small objects. Consequently, this affects the overall MOT performance. In fact, Ooi et~al. \cite{ooi2018multiple} used only detection inputs, which was not able to track any object in this video. It was already demonstrated that the use of only MOD objects as inputs for the MOT does not work well for this particularly challenging video. The good MOTP values obtained by Ooi et~al. \cite{ooi2018multiple} show that MOD BBs although not very reliable can give object locations that are sometimes more precise. 
\begin{table}[]
\centering
\caption{Comparison of the proposed method performance with state-of-the-art approaches. \textbf{Boldface} indicates best results, \textit{italic} indicates second best.}
\label{tab_comparison}
\begin{tabular}{ c | c c | c c| c c |c c |c c}
\hline 
    &   \multicolumn{2}{|c|}{Our method} & \multicolumn{2}{|c|}{IMOT} & \multicolumn{2}{|c|}{Urban Tracker} & \multicolumn{2}{|c}{MKCF} & \multicolumn{2}{|c}{Ooi et al.}\\
	&	MOTP	&	MOTA	&	MOTP	&	MOTA	&	MOTP	&	MOTA	&	MOTP	&	MOTA	&	MOTP	&	MOTA\\
\hline
Rouen	&	0.604	&	0.601	&	\textit{0.620}	&	\textit{0.670}	&	0.617	&	\textbf{0.696}	&	0.582	&	0.501 & \textbf{0.687} &  -0.188	\\
Sherb. 	&	0.582	&	\textit{0.595}	&	\textit{0.590}	&	\textbf{0.690}	&	0.576	&	0.404	&	0.553	&	0.317 & \textbf{0.749} & 0.027	\\
St-Marc	&	\textit{0.696} 	&	\textbf{0.654}	&	0.682	&	\textit{0.653}	&	0.691	&	0.638	&	0.652	&	0.463 & \textbf{0.723} &	-0.366\\
Rene-L.	&	\textbf{0.741}	&	0.230	&	\textit{0.705}	&	\textbf{0.613}	&	0.582	&	\textit{0.565}	&	0.531	&	0.334 & NA & NA	\\
\hline
\end{tabular}
\end{table}

\subsection{Discussion}
The integration of objects from IMOT and a MOD is proposed in order to better capture the objects of interest during the tracking process. It was expected that the combined inputs can complement each other, producing better inputs compared to the inputs produced individually from the different approaches. For instance, with the presence of fragmented objects from background subtraction that are difficult to group together, having a reference BB from the MOD that encompasses the whole object could be a useful indicator to improve the representation of the complete object. However, from the experiments, we have noticed the tendency of the MOD to generate large BBs that often include areas that do not belong to the object of interest. While in certain frames, it is helpful to have such BBs showing objects that are partially occluded, there are many occasions that such BBs include several objects of interest as one detection, especially for objects of small sizes such as pedestrians in urban traffic scene. 

This led to a difficulty of tracking them effectively as the input objects to the tracker are already merged as one whole object instead of distinct objects. In addition, there are cases where IMOT objects encompassed more than one object of interest that appeared on the scene as well due its origin of background subtraction. As an effort to mitigate these effects, we have imposed a stricter merging threshold to reduce the amount of incorrect fusion of objects. To distinguish the case between combining BBs of fragmented parts into one whole object, and the case of having multiple objects interacting in close proximity, we take into consideration the colour of IMOT objects to make the merging decision.

The excessive inclusion of areas that are not relevant may impact the tracking process as well. This is because the colour histogram will consider the background portion that was included in the BB for object description in the association cost for matching across frames, leading to possibly less accurate descriptions of the objects of interest. However, despite the effort to differentiate the two cases, some missed objects are still missed in the final tracking outputs because of the imperfect representation of some objects of interest that get fed into the tracker. The missed objects could be the result of MOD objects that are not paired with the available IMOT objects. Indeed, sometimes the MOD can detect object that IMOT cannot. 


\section{Conclusion}
In this paper, we presented a novel approach for fusing input objects from a multiclass object detector and an improved object extraction approach based on background subtraction for multiple object tracking. We use the integrated set of objects into a proposed MOT framework that associates objects across frames using spatial, colour and class label information to form trajectories in challenging urban traffic scenes. The prediction quality of unmatched objects in the MOT paradigm is evaluated to further improve the final tracking results. Results show that our method is competitive, but that it is very challenging to combine detections from multiple sources. First, they may not detect the same objects, and secondly, even if the same objects are detected, objects are not bounded in the same way. Our ablation study show that using class labels and their confidence can contribute positively to the data association cost function.


\subsubsection*{Acknowledgments}
This research is funded by FRQ-NT (Grant: 2016-PR-189250) and Polytechnique Montr\'eal PhD Fellowship. The Titan X used for this research was donated by the NVIDIA Corporation.

%
%
%
\bibliographystyle{splncs04}
\bibliography{mybibliography}

\begin{thebibliography}{10}
\providecommand{\url}[1]{\texttt{#1}}
\providecommand{\urlprefix}{URL }
\providecommand{\doi}[1]{https://doi.org/#1}

\bibitem{CSM2009Shalom}
Bar-Shalom, Y., Daum, F., Huang, J.: The probabilistic data association filter.
  IEEE Control Systems Magazine  \textbf{29}(6),  82--100 (Dec 2009)

\bibitem{beaupre2018improving}
Beaupr{\'e}, D.A., Bilodeau, G.A., Saunier, N.: Improving multiple object
  tracking with optical flow and edge preprocessing. arXiv preprint
  arXiv:1801.09646  (2018)

\bibitem{bewley2016simple}
Bewley, A., Ge, Z., Ott, L., Ramos, F., Upcroft, B.: Simple online and realtime
  tracking. In: 2016 IEEE International Conference on Image Processing (ICIP).
  pp. 3464--3468. IEEE (2016)

\bibitem{girshick2016region}
Girshick, R., Donahue, J., Darrell, T., Malik, J.: Region-based convolutional
  networks for accurate object detection and segmentation. IEEE transactions on
  pattern analysis and machine intelligence  \textbf{38}(1),  142--158 (2016)

\bibitem{henriques2015high}
Henriques, J.F., Caseiro, R., Martins, P., Batista, J.: High-speed tracking
  with kernelized correlation filters. IEEE transactions on pattern analysis
  and machine intelligence  \textbf{37}(3),  583--596 (2015)

\bibitem{jodoin2016tracking}
Jodoin, J.P., Bilodeau, G.A., Saunier, N.: Tracking all road users at
  multimodal urban traffic intersections. IEEE Transactions on Intelligent
  Transportation Systems  \textbf{17}(11),  3241--3251 (2016)

\bibitem{kuhn1955hungarian}
Kuhn, H.W.: The hungarian method for the assignment problem. Naval research
  logistics quarterly  \textbf{2}(1-2),  83--97 (1955)

\bibitem{luo2018mio}
Luo, Z., Branchaud-Charron, F., Lemaire, C., Konrad, J., Li, S., Mishra, A.,
  Achkar, A., Eichel, J., Jodoin, P.M.: Mio-tcd: A new benchmark dataset for
  vehicle classification and localization. IEEE Transactions on Image
  Processing  \textbf{27}(10),  5129--5141 (2018)

\bibitem{milan2016mot16}
Milan, A., Leal-Taix{\'e}, L., Reid, I., Roth, S., Schindler, K.: Mot16: A
  benchmark for multi-object tracking. arXiv preprint arXiv:1603.00831  (2016)

\bibitem{ooi2018multiple}
Ooi, H.L., Bilodeau, G.A., Saunier, N., Beaupr{\'e}, D.A.: Multiple object
  tracking in urban traffic scenes with a multiclass object detector. In:
  International Symposium on Visual Computing. pp. 727--736. Springer (2018)

\bibitem{CVPR2011Pirsiavash}
Pirsiavash, H., Ramanan, D., Fowlkes, C.C.: Globally-optimal greedy algorithms
  for tracking a variable number of objects. In: CVPR 2011. pp. 1201--1208
  (June 2011)

\bibitem{Rezatofighi2015}
Rezatofighi, S.H., Milan, A., Zhang, Z., Shi, Q., Dick, A., Reid, I.: Joint
  probabilistic data association revisited. In: 2015 IEEE International
  Conference on Computer Vision (ICCV). pp. 3047--3055 (Dec 2015)

\bibitem{saunier2006feature}
Saunier, N., Sayed, T.: A feature-based tracking algorithm for vehicles in
  intersections. In: The 3rd Canadian Conference on Computer and Robot Vision
  (CRV'06). pp. 59--59. IEEE (2006)

\bibitem{Shi1994}
Shi, J., Tomasi, C.: {Good features to track}. In: IEEE CVPR. pp. 593--600
  (1994)

\bibitem{yang2017multiple}
Yang, Y., Bilodeau, G.A.: Multiple object tracking with kernelized correlation
  filters in urban mixed traffic. In: 2017 14th Conference on Computer and
  Robot Vision (CRV). pp. 209--216. IEEE (2017)

\end{thebibliography}

\end{document}